\pdfoutput=1

\documentclass[11pt]{article}
\usepackage{graphicx}
\usepackage{multirow}
\usepackage[dvipsnames]{xcolor}
\usepackage{color}
\usepackage{booktabs}
\usepackage{amsmath}
\usepackage{fixltx2e}
\usepackage[]{acl}

\usepackage{times}
\usepackage{latexsym}

\usepackage[T1]{fontenc}

\usepackage[utf8]{inputenc}

\usepackage{microtype}

\title{Entity-based SpanCopy for Abstractive Summarization to Improve the Factual Consistency}

\author{Wen Xiao and Giuseppe Carenini\\
  Department of Computer Science \\
  University of British Columbia \\
  Vancouver, BC, Canada, V6T 1Z4 \\
  {\tt \{xiaowen3, carenini\}@cs.ubc.ca}}
\begin{document}
\maketitle
\begin{abstract}
Despite the success of recent abstractive summarizers on automatic evaluation metrics, %
the generated summaries still present %
factual inconsistencies with the source document. %
In this paper, we focus on entity-level factual inconsistency, i.e. reducing the mismatched entities between the generated summaries and the source documents. We therefore propose a novel entity-based SpanCopy mechanism, 
and explore its extension with a Global Relevance component 
. 
Experiment results on four summarization datasets show that SpanCopy can effectively improve the entity-level factual consistency with essentially no change in the word-level and entity-level saliency.
\footnote{The code is available at \url{https://github.com/Wendy-Xiao/Entity-based-SpanCopy}}
\end{abstract}

\section{Introduction}

Abstractive text summarization, %
the task to generating informative and fluent summaries of the given document(s), %
has attracted much attention in the NLP community. While early neural approaches focused more on designing customized architectures or training schema to better fit  %
the summarization task \cite{nallapati-etal-2016-abstractive,tan-etal-2017-abstractive,j.2018generating}, recent works have shown that %
generation models, %
pre-trained on large corpora \cite{lewis-etal-2020-bart,pegasus,t5}, generally have a better performance when fine-tuned on in-domain datasets.

However, %
even if these pre-trained\&fine-tuned generation models achieve state-of-the-art performance with respect to standard automatic evaluation metrics, e.g. ROUGE score\cite{lin-2004-rouge} and BERTScore\cite{bertscore}, the generated summaries still suffer from the problem of factual inconsistency, which means the generated summaries may not be factually consistent with the content expressed in the source documents \cite{kryscinski-etal-2020-evaluating}. %
Inconsistencies may exist either at the entity or the relation level \cite{nan-etal-2021-entity}. The former case is  when the summary mentions an \textit{entity} that does not appear in the source documents. The latter is %
when the summary does mention entities from the source documents, but expresses a \textit{relation} between them which is %
different than the one stated in the source documents.

In this paper, we focus on the entity-level inconsistency problem, %
i.e. to make the model generate summaries with less entities which do not appear in the source document(s) i.e., `hallucinated' entities. Note however, that hallucinated entities are not necessarily `unfaithful' or `wrong'\cite{hallucinated_but_factual}, so the goal is to reduce them without excluding entities that do appear in the reference summary i.e., without penalizing saliency.
Table~\ref{tab:example} shows an example of entity-level factual inconsistency from the XSum dataset. Although the content of the summary generated by the SOTA summarizer PEGASUS~\cite{pegasus} is roughly similar that of the ground-truth summary, it does not accurately summarize the original documents with the \textit{proper entities}. Specifically, it totally misses the entity `Royal Marine', which appears in both the source document and the reference summary,
and the entity `Hampshire' is `hallucinated', as it does not appear in the source document. Despite the fact that the city `Portsmouth' is located in `Hampshire' county, the entity itself is still an instance of factual inconsistency (i.e., an unnecessary generalization).
 \begin{table}[t!]
    \centering
    \begin{tabular}{p{0.95\linewidth}}
    \toprule

    \textbf{\textit{Entities in Source Doc:} } Royal Marine, Falklands, Portsmouth, Falklands War Memorial....\\
    \midrule
    \textbf{Ground Truth:} Plans to move a statue depicting a \textcolor{Green}{Royal Marine} in the \textcolor{Green}{Falklands} conflict away from \textcolor{Green}{Portsmouth} seafront have been criticised.\\
    \midrule
    \textbf{PEGASUS:} A campaign has been launched to keep a statue of a \textcolor{red}{Falklands War} marine in \textcolor{red}{Hampshire}.\\
    \midrule
    \textbf{SpanCopy:} A campaign to keep a statue of a \textcolor{Green}{Royal Marine} marching across the \textcolor{Green}{Falklands} in \textcolor{Green}{Portsmouth} has been launched.\\
    \midrule
    \textbf{SpanCopy + GR:}  A statue of a \textcolor{Green}{Royal Marine} marching across the \textcolor{Green}{Falklands} during the \textcolor{RoyalBlue}{Falklands War Memorial} should remain in its current location, campaigners have said.\\
    \bottomrule
    \end{tabular}
    \caption{An example of entity-level factual inconsistency from the XSum dataset. The summary generated by PEGASUS totally missed one entity (Royal Marine)
    and one entity indicates a larger area than the correct one (Hampshire).}
    \label{tab:example}
\end{table}

Prior work~\cite{dong-etal-2020-multi,constraining_beam_search} mainly address the entity-level inconsistency problem in the post-processing stage. However, those methods either requires additional sophisticated models, e.g.  \citet{dong-etal-2020-multi} uses a pre-trained QA model to `revise' the generated summaries, or being built on arguably brittle heuristics~\cite{constraining_beam_search}
. Recent work~\cite{nan-etal-2021-entity} proposes two ways to directly improve the end-to-end summarization model, either by training with an auxiliary task, which is to recognize the summary-worthy entities in the source document using the hidden states from the encoder, or jointly generating the entities and the summaries, i.e. generating a chain of entities in the summary followed by the summary. Yet, both methods do not explicitly encourage the model to generate the summaries with more valuable entities, as both of them aim to guide the model to detect the summary-worthy entities without any changes in the summary generation process. Instead, aiming for a lean and modular solution, we propose the SpanCopy Mechanism to explicitly copy the matched entities\footnote{We particularly focus on the Named Entities in this paper, but our method can be easily applied to any kinds of span or entities.} from the source documents when generating the summaries. One key advantage of our proposal is that %
it can be easily integrated into any pre-trained generative sequence-to-sequence model.

Since often only a few %
of the entities in the source documents can  be included in the summary, which we call `summary-worthy entities', we also explore an additional  Global Relevance component to better recognize the summary-worthy entities by automatically generating a prior distribution over all the entities in the source documents. 

We test our proposal on four summarization datasets in the news and scientific paper domain, comparing it with the SOTA PEGASUS system~\cite{pegasus}. In a first set of experiments, as a sanity check, we assess our models on arguably easier subsets of these  datasets, where all the entities in the reference summaries belong to the source document. In these cases, SpanCopy should definitely dominate PEGASUS, which is confirmed by the results.
In a  second set of experiments, we fine-tune and test on the full datasets. On this realistic and more challenging task, we find that 
SpanCopy (without Global Relevance) can strongly improve the entity-level factual consistency ($+2.28$) on average across datasets, with essentially no change in saliency ($-0.06$).

\begin{figure*}[th]
    \centering
    \includegraphics[width=\linewidth]{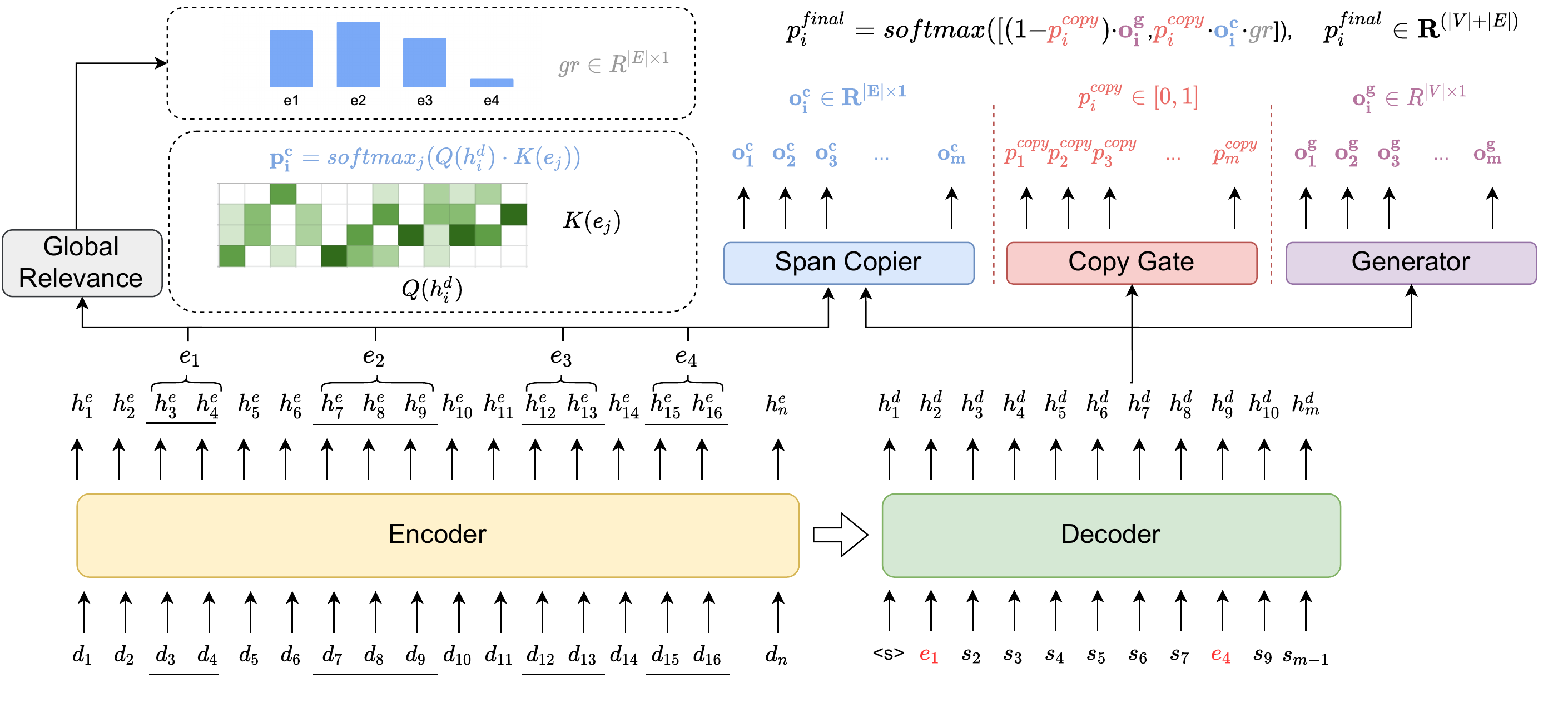}
    \caption{Structure of the model with Entity-based SpanCopy Mechanism, with five components: \textcolor{Dandelion}{Encoder}, \textcolor{Green}{Decoder}, \textcolor{RoyalBlue}{Span Copier}, \textcolor{Red}{Copy Gate} and \textcolor{Purple}{Generator}. The upper left bar plot shows the \textcolor{Gray}{Global Relevance} component, predicting the prior probability of all the entities $\{e_1,e_2,e_3,e_4\}$ to be copied to the summary.}
    \label{fig:model}
\end{figure*}

\section{Related Work}
\subsection{Abstractive Summarization} 
Early neural abstractive summarization models~\cite{nallapati-etal-2016-abstractive,paulus2018a,see-etal-2017-get} are mainly sequence-to-sequence models based on different variants of RNN, e.g. LSTM or GRU, with additional components targeting different properties of the summaries, like redundancy~\cite{tan-etal-2017-abstractive} and coverage~\cite{see-etal-2017-get}. %
However, all the recurrent models suffer from  serious weakness like long-term memory loss, and requiring excessive time to train.

To tackle these problems, researchers in the area of abstractive summarization started to use %
attention-based transformer models %
\cite{liu-lapata-2019-hierarchical,liu-lapata-2019-text}; recently reaching SOTA performance when pre-trained generative transformers are applied to the %
task, e.g. BART~\cite{lewis-etal-2020-bart}, PEGASUS~\cite{pegasus} and PRIMERA~\cite{primera}. 
The SpanCopy mechanism we propose in this paper can be advantageously injected into any pre-trained models. %
\subsection{Factual Consistency}
Despite the large improvements with respect to automatic evaluation metrics, recent studies~\cite{Cao_Wei_Li_Li_2018,kryscinski-etal-2020-evaluating} show that around 30\% of the summaries generated by the SOTA summarization models contain factual inconsistencies. %
Ideally, the assessment of factual consistency should rely on human annotations~\cite{maynez-etal-2020-faithfulness}, but these are costly, time consuming and lack a unified standard. Thus promising automatic evaluation metrics for factual consistencies of generated summaries have been %
explored in recent years. 
To assess relation-level factual consistency two kinds of metrics have been proposed: one based on classification ~\cite{kryscinski-etal-2020-evaluating}, and one based on Question-Answering~\cite{maynez-etal-2020-faithfulness,durmus-etal-2020-feqa}. For entity-level factual consistency, the focus of this paper, 
\citet{nan-etal-2021-entity} propose a simple but effective evaluation metric, %
based on the matched named entities in both generated and ground-truth summaries. In our work, we use %
such metric to evaluate whether the generated summaries are consistent with both the source documents and the reference summaries at the  entity-level.

\subsection{Copy Mechanism}
\citet{see-etal-2017-get} first apply pointer-generator network in an abstractive summarization model, which facilitates copying words from the source documents by pointing, i.e., generating a distribution of probabilities to copy each word from the source. Following their work, \citet{bi-etal-2020-palm} propose PALM, in which the copy mechanism is applied on top of the transformer model, and with a novel pre-training schema, the model achieves SOTA on several generative tasks, such as abstractive summarization and generative QA. More recently,  \citet{li-etal-2021-learn} further explores how to make use of the copy history to predict the copy distribution for the current step. However, all the aforementioned works focus on copying at the  word level, which tends to be sparse and noisy. Instead, we aim to train the model to copy spans of text i.e., the named entities, in this paper.

Admittedly, some previous work has also investigated span-based copy mechanisms. Yet, those models either predict the start and end indices of a span~\cite{Zhou_Yang_Wei_Zhou_2018}, or predict the BIO labels %
for each token ~\cite{liu-etal-2021-biocopy}. Even if such strategies can copy any kinds of spans (clauses, n-grams, entities, phrases or longest common sequence) from the source document, they may introduce unnecessary noise and break the coherence of the generated text. In this work, we focus on copying the spans of the Named Entities, extracted by a high-quality NER tool, aiming to improve factual consistency of the generated  summary without negatively affecting saliency.

\section{Our SpanCopy Method}

\subsection{Transformer-based %
Summarizers}
Typically, transformer-based %
summarization\cite{lewis-etal-2020-bart,pegasus} %
consists of two steps (i) The %
\textbf{Encoding Step} (by the Encoder shown in yellow in Fig.\ref{fig:model}), which encodes the source input(s) into an hidden space; (ii) the %
\textbf{Decoding Step}, which computes a probability distributions on the output vocabulary to generate each token of the resulting summary. %
In this paper, to better describe our methods in the context of a generic %
summarization models, we %
split the Decoding process into two components, the Decoder itself (shown in green in Fig.\ref{fig:model}), which outputs the representations of predicted tokens, and the Generator (shown in purple in Fig.\ref{fig:model}), an MLP layer mapping the representations to the final probability distribution on the output vocabulary. 

More formally, for a document with n tokens $D=\{t^d_1,t^d_2,...,t^d_n\}$, and the corresponding summary with m tokens, $S=\{t^s_1,t^s_2,...,t^s_m\}$, the output of the Encoder is a sequence of hidden states of all the tokens, i.e. $\{h^e_1,h^e_2,...,h^e_n\}$. And then the Decoder predicts a sequence of vector, $\{h^d_1,h^d_2,...,h^d_m\}$, representing the tokens to be predicted. Finally, the Generator maps those vectors to the distributions over the vocabulary, i.e. $\{\mathbf{p}_1,\mathbf{p}_2,...,\mathbf{p}_m\}$, where $\mathbf{p}_i \in R^{|V|}$.

\subsection{SpanCopy Mechanism} 
A key problem with generic sequence-to-sequence transformer-based summarizers is that the decoding step is prone to generate factual inconsistencies, i.e. the model may make up entities or relations that are not entailed by the source documents. To address entity-level factual inconsistency, we introduce in the Decoding Step the SpanCopy mechanism, which can be conveniently plugged into any pre-trained models. Specifically, we first identify and match the entities in both source document and summary, and then instead of generating the entire summary word by word, we add an additional Span Copier to directly copy entities from the source document, with a Copy Gate predicting the likelihood of whether the model should generate the current token from the vocabulary or directly copy an entity from the source document.

\paragraph{Span Copier} (shown in blue in Fig.\ref{fig:model}) is an attention module over all the entities in the input document. Suppose there are $|E|$ entities in the input document, with each entity $j$ being a span over tokens $[d_{j_s},d_{j_e}]$, then the entities can be simply represented as $e_j=\mathbf{avg}([h^e_{j_s}:h^e_{j_e}])$, where $h^e_i$ represents the output of the encoder for each token $d_i$. At each decoding step $i$, we compute the logit vector of copying each entity at the current step as:
\begin{equation}
    \mathbf{o^c_i} = Q(h_i^d)\cdot K(e_j),\mathbf{o^c_i} \in \mathbf{R}^{|E|}
\end{equation}
indicating how likely it is to copy the entities from the source document at each step. Notice that to better balance the numeric difference caused by the size of selection space ($|V|$ and $|E|$), we generate and combine the raw logit vectors\footnote{The vector of raw (non-normalized) predictions that the classification model generates} from the Span Copier and Generator, and take softmax over the combined space to get the final probability.

\paragraph{Copy Gate} (shown in red in Fig.\ref{fig:model}) is a classifier to map the hidden states to a singular value, i.e. 
\begin{equation}
p^{copy}_i= \sigma(MLP(h^d_i)), p^{copy}_i \in [0,1]
\end{equation}
which indicates the probability of copying an entity at each step. On the contrary, $1-p^{copy}_i$ represent the probability of generating a token from the vocabulary at step $i$.

Then the final probability, combining both generation over the vocabulary and the copy mechanism over the entity space, is computed as  
\begin{eqnarray}
\label{eq: pfinal}
\mathbf{p^{final}_i} =softmax( [(1-p^{copy}_i)\cdot\mathbf{o^g_i},p^{copy}_i\cdot\mathbf{o^c_i}])
\end{eqnarray}
with $\mathbf{p^{final}_i} \in \mathbf{R}^{(|V|+|E|)}$, where $\mathbf{o^g_i} \in \mathbf{R}^{(|V|)}$ is the logit vector of token generation and $\mathbf{o^c_i} \in \mathbf{R}^{(|E|)}$ is the logit vector of entity copying. As a result, the first $|V|$ dimensions of the final probability represent the probability of generating all the tokens from the vocabulary, while the following $|E|$ dimensions contain the probabilities of copying the entities from the source document.

Note that the input of the original Decoder in the transformer model at each step is the embedding of the previous token (which is the ground-truth one during training, and the predicted one for inference), but a span of text longer than $1$ does not naturally have an embedding to match. We simply use the average of the embedding of all the tokens in the entity, following previous work using average embedding to represent a span of text \cite{xiao-carenini-2019-extractive}.
\subsection{Loss}

We use the standard loss for  abstractive summarization, i.e. the cross entropy loss between the predicted probability and the ground truth labels,
\begin{equation}
    L_1 = \sum_i L_{s}(\mathbf{p^{final}_i},t_i)
\end{equation}
However, notice that, since the predicted probability distribution is over the combined space of vocabulary size and entity size ($\mathbf{p^{final}_i} \in \mathbf{R}^{|V|+|E|}$), the corresponding ground truth labels can be either indices of words to be generated from the vocabulary, or the indices of entities to be copied from the source document, i.e. $t_i \in [0,|V|+|E|]$.  Specifically, if $t_i<|V|$, then the $t_i$-th token should be generated, and if $t_i>|V|$, the $(t_i-|V|)$-th entity should be copied from the source document.   

\begin{table*}[th!]
    \centering
    \resizebox{0.85\linewidth}{!}{\begin{tabular}{l|rrrrr|rrrrr}
    \toprule
    
    \multirow{2}{*}{Dataset}       &\multicolumn{5}{c|}{Original}&\multicolumn{5}{c}{Filtered}\\
    & $L_{doc}$& $L_{summ}$&$N_{doc}$& $N_{summ}$ &$src_p(gt)$&$L_{doc}$& $L_{summ}$&$N_{doc}$& $N_{summ}$ &$src_p(gt)$ \\
    \midrule
    CNNDM     &690.9&52.0&42.8&5.9&80.41 &671.9&47.1&39.4&4.4&100\\
    XSum&373.8&21.1&27.9&2.7&39.85&483.4&20.6&31.6&1.9&100\\
    Pubmed&3049.0&202.4&71.1&6.4&70.93&3165.4&178.5&69.9&3.4&100 \\
    arXiv& 6033.3&271.5&157.5&6.0&39.12&6478.9&164.1&161.9&2.3&100\\
    \bottomrule
    \end{tabular}}
    \caption{Statistics of all the datasets (original/filtered), on the lengths ($L_{doc}$,$L_{summ}$) and number of entities ($N_{doc}$, $N_{summ}$) in the source documents and ground truth summaries, as well as $src_p(gt)$, the entity level source-precision of the ground-truth summary.} 
    \label{tab:stats_dataset}
\end{table*}
\subsection{SpanCopy with Global Relevance}
Among all the entities in the source documents, there are only a few summary-worthy entities that should be copied into the summary (e.g. around $10\%$ in CNNDM and $1.5\%$ in arXiv).
To make the model better recognize such summary-worthy entities, %
we explore a Global Relevance (GR) component, which takes all the entities in the source document as inputs, and predicts how likely each entity is to appear in the final summary. We use the generated `entity likelihood' as a prior distribution for the Span Copier component, with GR also trained as an auxiliary task.

\paragraph{Global Relevance} is a classifier mapping the hidden state of a source document entity into a %
value within $[0,1]$, indicating the probability %
that such entity should be included in the summary. 
\begin{equation}
    \mathbf{gr}= \sigma(MLP(\mathbf{e})), \mathbf{gr}\in \mathbf{R}^{|E|}
\end{equation}
Then $p^{final}_i$ %
in Eq.\ref{eq: pfinal} is updated with $gr$ as 
\begin{equation}
\begin{split}
\mathbf{p^{final}_i} =softmax(&  [(1-p^{copy}_i)\cdot\mathbf{o^g_i}\\
& ,p^{copy}_i\cdot\mathbf{o^c_i}\cdot\mathbf{gr}])
\end{split}
\end{equation}

\paragraph{New Loss} As an auxiliary task, we also train the model with the ground-truth %
GR labels to make it more accurate. Specifically, the label $y^{gr}_i=1$ if  the $i$-th entity in the input document is included in the ground truth summary. Then we update the loss function with $L_{gr}$ balanced by $\beta$:
\begin{equation}
\begin{split}
    L_2 = & (1-\beta)\sum_i L_{s}(\mathbf{p^{final}_i},t_i)\\
    & +\beta \sum_j L_{gr}(gr_j,y^{gr}_j)
\end{split}
\end{equation}

\section{Experiments and Analysis}
   \begin{table}[t!]
    \centering
    \resizebox{\linewidth}{!}{\begin{tabular}{ccc}
    \toprule
    Dataset     & \# Data (original) & \# Data (filtered) \\
    \midrule
    CNNDM     &287,113/13,368/13,368&105,847/4,490/3,903\\
    XSum&204,017/11,327/11,333&42,481/2,349/2,412\\
    Pubmed&119,924/6,633/6,658&32,123/1,797/1,772\\
    arXiv&202,914/6,436/6,440&66,360/2,365/2,324\\
    \bottomrule
    \end{tabular}}
    \caption{Number of data examples in all the datasets (original v.s. filtered).}
    \label{tab:number_data}
\end{table}
\subsection{Settings}
SpanCopy %
can be plugged into any pre-trained generation model. In this paper, we use PEGASUS\cite{pegasus} as our base model, since it has %
delivered top performance
on multiple summarization datasets. We recognize %
named entities with an off-the-shelf NER tool\footnote{https://spacy.io/}. The balance factor $\beta$ of GR %
is set by grid search on small subsets  of each dataset (2k for training and 200 for validation).
\subsection{Evaluation Metrics}
To %
evaluate the saliency and entity-level factual consistency of the generated summaries, we apply the following metrics:
\paragraph{Saliency metrics} assess the similarity of the generated summary with the reference summary.

\textit{ROUGE scores} \cite{lin-2004-rouge}  measure the n-gram overlaps between generated and ground truth summaries. We apply the %
metrics R-1, R-2 and R-L.

\textit{Summary-precision, -recall and -f1} ($sum_p$, $sum_r$ and $sum_f$)  \cite{nan-etal-2021-entity}  measure the precision/recall/f1 score of the matched entities in the generated summaries and the reference summaries. we use  $NE(S_{ref})$ and $NE(S_{gen})$ to represent the named entities in the  reference summaries and generated summaries, respectively.
    \begin{eqnarray*}
    sum_p&=&|NE(S_{ref})\cap NE(S_{gen})|/|NE(S_{gen})|\\
        sum_r&=&|NE(S_{ref})\cap NE(S_{gen})|/|NE(S_{ref})|\\
        sum_f&=&2*(sum_p+sum_r)/sum_p*sum_r
    \end{eqnarray*}
    These three metrics measure the entity-level saliency  of the generated summaries, i.e. recognizing how many copied (and generated) entities are salient, and should be included in the summary. 
    
    \begin{table*}[t!]
    \centering
    \footnotesize
    \begin{tabular}{l|ccc|ccc|c}
    \toprule
    \multirow{2}{*}{Model}  &\multicolumn{3}{c}{ROUGE} &\multicolumn{3}{c}{Entity(Summ)}& \multicolumn{1}{c}{Entity(Doc)}\\

    &R-1&R-2&R-L&$sum_r$ &$sum_p$&$sum_f$&$src_p$\\
    
    \midrule
    \multicolumn{8}{c}{CNNDM Filtered}\\
    \midrule
    PEGASUS & \textcolor{Red}{44.70}&\textcolor{Red}{22.23}&\textcolor{Red}{32.52}&\textcolor{Red}{50.80}&\textcolor{Red}{45.32}&\textcolor{Red}{45.03}&\textcolor{Red}{92.85}\\
    SpanCopy&45.46&23.12&33.48&53.08&48.63&47.86&94.64\\
    SpanCopy+GR&\textcolor{Green}{45.74}&\textcolor{Green}{23.44}&\textcolor{Green}{33.67}&\textcolor{Green}{54.61}&48.27&\textcolor{Green}{48.36}&\textcolor{Green}{95.02}\\
    \midrule 
    \multicolumn{8}{c}{XSum Filtered}\\
    \midrule
   PEGASUS&\textcolor{Red}{43.01}&\textcolor{Red}{19.00}&\textcolor{Red}{34.01}&\textcolor{Red}{59.14}&\textcolor{Red}{54.94}&\textcolor{Red}{54.68}&\textcolor{Red}{77.32}\\
    SpanCopy&\textcolor{Green}{44.23}&\textcolor{Green}{19.90}&\textcolor{Green}{35.50}&\textcolor{Green}{61.34}&59.15&58.16&\textcolor{Green}{84.30}\\
    SpanCopy+GR&43.78&19.12&34.97&60.69&\textcolor{Green}{60.50}&\textcolor{Green}{58.36}&83.75\\
    \midrule
    \multicolumn{8}{c}{Pubmed Filtered}\\
    \midrule
   PEGASUS&\textcolor{Red}{46.99}&\textcolor{Red}{21.46}&\textcolor{Red}{42.57}&\textcolor{Green}{42.63}&\textcolor{Red}{33.28}&\textcolor{Red}{33.16}&\textcolor{Red}{73.59}\\
    SpanCopy&47.82&\textcolor{Green}{22.34}&43.43&\textcolor{red}{41.58}&34.12&33.44&73.74\\
    SpanCopy+GR&\textcolor{Green}{48.04}&22.18&\textcolor{Green}{43.56}&42.11&\textcolor{Green}{36.21}&\textcolor{Green}{34.86}&\textcolor{Green}{74.15}\\
    \midrule
    \multicolumn{8}{c}{arXiv Filtered}\\
    \midrule
   PEGASUS&\textcolor{red}{46.23}&\textcolor{red}{18.02}&\textcolor{red}{41.02}&37.65&\textcolor{red}{35.98}&33.48&68.13\\
    SpanCopy&46.36&\textcolor{Green}{18.29}&41.23&\textcolor{Green}{39.50}&\textcolor{Green}{37.61}&\textcolor{Green}{34.95}&\textcolor{Green}{72.12}\\
    SpanCopy+GR&\textcolor{Green}{46.56}&18.27&\textcolor{Green}{41.34}&\textcolor{red}{35.38}&36.11&\textcolor{red}{32.76}&\textcolor{red}{67.56}\\
    \bottomrule
    \end{tabular}
    \caption{Result of our models and the compared backbone model (PEGASUS) on the filtered datasets. ROUGE score and Entity(Summ) are mainly used to measure the word-level saliency and entity-level saliency, respectively. Entity(Doc) is used to measure the entity-level factual consistency. \textcolor{red}{Red} represents the lowest among all the three models, while \textcolor{Green}{Green} represents the highest.}
    \label{tab:result_filtered}
    \vspace{-3mm}
\end{table*}
\paragraph{Entity-level factual consistency metric:} measures the named entity matching between the generated summaries and the source documents. 
\cite{nan-etal-2021-entity} 
With $NE(D)$ and $NE(S_{gen})$ representing the named entities in the source document and generated summaries, respectively, 
\textit{Source-precision}($src_p$) measures how many entities in the generated summaries are from the source documents, i.e. $src_p=|NE(D)\cap NE(S_{gen})|/|NE(S_{gen})|$. It is an evaluation metric for entity-level factual consistency, as it directly measures how consistent the generated summaries are with the source.

\subsection{Datasets}
We test and compare our SpanCopy model with the original PEGASUS on four datasets, in the domains of news (CNNDM\cite{nallapati-etal-2016-abstractive}, XSum\cite{narayan-etal-2018-dont}) and scientific papers (Pubmed and  arXiv\cite{cohan-etal-2018-discourse}). As a sanity check, we initially assess our models on subsets of these datasets, where all the entities in the reference summaries belong to the source document (we call these filtered datasets). In these cases ($src_p(gt)=1$), Spam Copy and GR  should  dominate PEGASUS, because  by design they tend to generate entities from the source document.
We compare the size of filtered and original datsets in Table~\ref{tab:number_data}. 

The statistics of the filtered and original datasets, on the lengths and number of entities in the document and summaries, can be found in Table~\ref{tab:stats_dataset}. $src_p(gt)$ measures the entity-level factual consistency between the source document and the ground-truth summary, with lower value meaning that there are more novel entities in the ground-truth summaries. %
The table shows that the datasets in the news domain have higher density of the entities with respect to the lengths (number of words) of both documents and ground-truth summaries, i.e. $N_{doc}/L_{doc}$ and $N_{summ}/L_{summ}$ are larger for the news articles. a possible explanation  is that news articles tend to describe an event or a story, which may contain more names of people, organizations, locations, etc., as well as  dates. Interestingly, CNNDM and Pubmed contain less novel than the other two datasets (with higher $src_p(gt)$), something that the proposed SpanCopy mechanism may benefit from. %
Comparing the filtered datasets with the original ones, we can see that the number of entities in the summaries drops for all the datasets, especially for arXiv, as the more entities in the summary, the less likely they can be all matched to the source documents.

\subsection{Results and Analysis}

\begin{table*}[th!]
    \centering
    \footnotesize
    \begin{tabular}{l|ccc|ccc|c}
    \toprule
    \multirow{2}{*}{Model}  &\multicolumn{3}{c}{ROUGE} &\multicolumn{3}{c}{Entity(Summ)}& \multicolumn{1}{c}{Entity(Doc)}\\

    &R-1&R-2&R-L&$sum_r$&$sum_p$&$sum_f$&$src_p$\\
    
    \midrule
    \multicolumn{8}{c}{CNNDM}\\
    \midrule
    PEGASUS & \textcolor{Green}{44.62}&20.82&31.05&\textcolor{Green}{46.87}&\textcolor{red}{42.25}&\textcolor{Green}{42.29}&\textcolor{red}{89.92}\\
    SpanCopy&44.19&\textcolor{Green}{20.86}&\textcolor{Green}{31.19}&43.15&\textcolor{Green}{43.87}&41.25&\textcolor{Green}{91.89}\\
    SpanCopy+GR&\textcolor{red}{44.16}&\textcolor{red}{20.61}&\textcolor{red}{30.97}&\textcolor{red}{42.72}&43.34&\textcolor{red}{40.79}&91.31\\
    \midrule 
    \multicolumn{8}{c}{XSum}\\
    \midrule
    PEGASUS&\textcolor{Green}{46.65}&\textcolor{Green}{23.47}&\textcolor{Green}{38.67}&\textcolor{Green}{41.09}&\textcolor{Green}{44.43}&\textcolor{Green}{40.96}&\textcolor{red}{41.23}\\
    SpanCopy&46.23&22.76&37.96&\textcolor{red}{39.90}&42.97&39.70&41.89\\
    SpanCopy+GR&\textcolor{red}{46.02}&\textcolor{red}{22.36}&\textcolor{red}{37.58}&40.12&\textcolor{red}{42.66}&\textcolor{red}{39.67}&\textcolor{Green}{42.79}\\
    \midrule
    \multicolumn{8}{c}{Pubmed}\\
    \midrule
    PEGASUS &\textcolor{red}{46.11}&\textcolor{red}{19.43}&\textcolor{red}{41.22}&\textcolor{red}{22.12}&\textcolor{red}{24.81}&\textcolor{red}{20.61}&67.03\\
    SpanCopy&46.21&\textcolor{Green}{19.86}&41.51&\textcolor{Green}{23.47}&25.10&21.29&\textcolor{Green}{68.91}\\
    SpanCopy+GR&\textcolor{Green}{46.27}&19.82&\textcolor{Green}{41.59}&23.34&\textcolor{Green}{25.29}&\textcolor{Green}{21.39}&\textcolor{red}{66.91}\\
    \midrule
    \multicolumn{8}{c}{arXiv}\\
    \midrule
    PEGASUS&\textcolor{Green}{44.23}&\textcolor{red}{16.55}&\textcolor{Green}{39.15}&\textcolor{Green}{20.98}&\textcolor{red}{25.42}&\textcolor{Green}{20.56}&\textcolor{red}{52.70}\\
    SpanCopy&44.05&16.76&\textcolor{red}{38.91}&20.65&25.46&20.39&\textcolor{Green}{56.88}\\
    SpanCopy+GR&\textcolor{red}{44.00}&\textcolor{Green}{16.87}&38.92&\textcolor{red}{20.01}&\textcolor{Green}{25.75}&\textcolor{red}{20.15}&54.21\\
    \bottomrule
    \end{tabular}
    \caption{Result of our models and the compared backbone model (PEGASUS) on the unfiltered datasets. See Table~\ref{tab:result_filtered} for the details of the columns.}
    \label{tab:result_unfiltered}
    \vspace{-3mm}
\end{table*}

\begin{table}[th!]
    \centering
    \footnotesize
    \begin{tabular}{l|r|r|r}
    Model     & $R_{avg}$&$sum_f$&$src_p$ \\
    \midrule
    \multicolumn{4}{c}{CNNDM}\\
    \midrule
     SpanCopy    & -0.08& -1.04&+1.97\\
     SpanCopy+GR&-0.25&-1.50&+1.39\\
    \midrule
    \multicolumn{4}{c}{XSum}\\
    \midrule
    SpanCopy    & -0.61& -1.26&+0.66\\
     SpanCopy+GR&-0.94&-1.29&+2.16\\
    \midrule
    \multicolumn{4}{c}{Pubmed}\\
    \midrule
    SpanCopy    & +0.27& +0.68&+1.88\\
     SpanCopy+GR&+0.31&+0.78&-0.12\\
    \midrule
    \multicolumn{4}{c}{arXiv}\\
    \midrule
    SpanCopy    & +0.20& +1.47&+3.99\\
     SpanCopy+GR&+0.30&-0.72&-0.57\\
     \midrule
    \multicolumn{4}{c}{\textbf{Overall} (avg. across all datasets)}\\
    \midrule
    SpanCopy    & -0.06& -0.04&+2.13\\
     SpanCopy+GR&-0.15& -0.68&+0.72\\
     \bottomrule
    \end{tabular}
    \caption{The relative ROUGE score (avg of R-1, R-2 and R-L), the entity-level summary-f1 and source-precision of our models, compared with the PEGASUS model on the four datasets (original). The last block shows the overall performance for all the datasets. }
    \label{tab:relative_results}
    \vspace{-3mm}
\end{table}

The results on the filtered and original datasets are shown in Table~\ref{tab:result_filtered} and Table~\ref{tab:result_unfiltered}. 

\paragraph{Filtered Datasets}  We first evaluate our models, with the backbone model, PEGASUS on the filtered datasets, which is an easier task, and the results can be found in Table~\ref{tab:result_filtered}. All the models are fine-tuned and tested on the filtered datasets. Since we only keep the examples with all the entities in the summaries being matched with the entities in the source documents, the theoretical ceiling of $src_p$ is 100.  Comparing SpanCopy and PEGASUS, SpanCopy performs better than PEGASUS regarding both saliency and entity-level factual consistency. Plausibly, this is because all the entities in the ground-truth summary can be copied from the source document, in which case the SpanCopy mechanism can better learn to copy. The SpanCopy model with the %
GR component performs better regarding the entity-level saliency on three out of all the four datasets. On arXiv, the performance of SpanCopy with the GR component regarding both entity-level saliency and factual consistency is quite low.  One likely reason might be that it is a rather difficult task to identify the salient entities in the arxiv dataset, as there is a large amount of entities in the source documents, but only very few entities are summary-worthy ($164.1$ v.s. $2.3$ as shown in Table~\ref{tab:stats_dataset}), which might bring in excessive noise.
\begin{table*}[t]
    \centering
    \footnotesize
    \begin{tabular}{p{0.95\linewidth}}
    \toprule
    \textbf{\textit{Entities in the Source Document:}} Yemen(0.28), Americans(0.25), Saudi Arabia(0.23), the State Department(0.23), CNN(0.20),..., U.S.(0.15), ...\\
    \midrule
    \textbf{Ground-truth Summary:} No official way out for \textcolor{Green}{Americans} stranded amid fighting in \textcolor{Green}{Yemen}. \textcolor{Green}{U.S.} \textcolor{Green}{Deputy Chief of Mission} says situation is very dangerous so no mass evacuation is planned .
\\
    \midrule
    \textbf{PEGASUS:}   \textcolor{RoyalBlue}{CNN}'s \textcolor{red}{Ivan Watson} joins a mother and her grandchildren waiting to be evacuated from \textcolor{Green}{Yemen}. \textcolor{RoyalBlue}{The State Department} has said it is too risky to evacuate Americans from the area. \textcolor{red}{Watson} meets \textcolor{Green}{Americans} who were on a \textcolor{RoyalBlue}{CNN} ship that docked at a \textcolor{red}{Yemeni} port.
\\
    \midrule
    \textbf{SpanCopy:} \textcolor{Green}{Dozens} of \textcolor{Green}{Americans} are trapped in \textcolor{Green}{Yemen}. The \textcolor{Green}{U.S.} has said it is too dangerous to evacuate \textcolor{Green}{Americans}.\\
    \midrule
    \textbf{SpanCopy+GR: } The \textcolor{Green}{U.S.} has said it is too dangerous to evacuate \textcolor{Green}{Americans} from \textcolor{Green}{Yemen}. \textcolor{RoyalBlue}{The State Department} said it is too risky to conduct an evacuation of citizens. A group of \textcolor{Green}{U.S.} organizations have filed a lawsuit against the government's stance on evacuations.\\
    \bottomrule
    \end{tabular}
    
    \caption{Example of the entity-level factual inconsistency, taken from the CNNDM dataset. The first block shows the entities in the source document with high GR scores (shown in parenthesis) from the SpanCopy + GR model. }
    \vspace{-3mm}
    \label{tab:example2}
\end{table*}
\paragraph{Original Datasets} %
In a second set of experiments, we fine-tune and test on the full/original datasets. On this realistic and more challenging task results are encouraging. As shown in Table~\ref{tab:result_unfiltered}, when the SpanCopy model is compared to PEGASUS, it improves the factual consistency of generated summaries with the source documents ($src_p$) on all the datasets, maintaining a very similar performance %
on the %
saliency metrics, i.e. ROUGE and entity-level saliency.  Comparing across the four datasets, SpanCopy outperforms PEGASUS on both the saliency and factual consistency metrics on the Pubmed dataset. For better comparison, we show the relative gains/loss regarding PEGASUS on all the datasets, as well as the overall average results in Table~\ref{tab:relative_results}. It is clear that the SpanCopy model performs much better regarding entity-level factual consistency ($+2.13$) with essentially no change in saliency ($-0.06$ on average ROUGE and $-0.04$ on entity-level saliency). Admittedly, despite the success of the GR component  on the filtered datasets on both word-level and entity-level saliency, it fails to deliver any gain on the original datasets. A plausible explanation is that  GR makes the model focus excessively on the entities in the source document, therefore penalizing generation of  new, potentially summary-worthy, entities.

Comparing the entity-level 
factual consistency on the filtered datasets and the original datasets, the filtered datasets always have higher $src_p$ than the original ones, and the gain is especially larger on the XSum and arXiv datasets, as both of them contain more entity-level hallucinations in the original datasets. Remarkably, the performance gain of the SpanCopy model over PEGASUS on the filtered XSum dataset is much larger on the original XSum datasets ($7.98$ v.s. $0.66$) , which might be because original XSum is more abstractive, the entity-level guidance is especially helpful for the abstractive examples with consistent entities in the summary.

\subsection{Qualitative Analysis}

For illustration, we examine a real example from the CNNDM dataset in Table~\ref{tab:example2}, which is a news article on the evacuation of Americans during the time of the crossfire of  warring parties in Yemen. While all of the three system generated summaries %
are able to capture the main statement that `it's too dangerous to evacuate the Americans', the person `Ivan Watson' mentioned by PEGASUS's summary does not exist in the source document, i.e., it is an `hallucinated' entity. Most likely, PEGASUS is generating  such hallucination because `Ivan Watson' is a senior CNN correspondent several time associated with Yemen in other news article in the training set, %
and the model automatically `picked the entity from the memory' to generate the summary without tightly adhering to the given document. In contrast, both of our models do not contain entities that are not in the source document, as the SpanCopy mechanism tend to guide the model to use more the entities in the source document. In addition, with the GR component, although the generated summary contains more matched entities with the source document, it pushes the model too far towards copying  entities which are not salient (e.g. \textit{The State Department}). 

\section{Conclusion and Future Work}
\vspace{-1mm}
In this paper, we tackle the problem of entity-level factual consistency for abstractive summarization, by guiding the model to directly copy the summary-worthy entities from the source document through the novel SpanCopy mechanism (with the optional GR component), which can be integrated into any transformer-based generative frameworks.
By running a sanity check on arguably easier subsets of four diverse summarization datasets, %
SpanCopy with GR is confirmed to perform better on both entity-level factual consistency and saliency. More tellingly, the experiments on the original test sets show that the SpanCopy mechanism can effectively improve the entity-level factual consistency with essentially no change in the word-level and token-level saliency.
In the future, we plan to extend %
our approach towards controllable generation with given entities. Specifically, instead of using the learnt GR scores, the model could generate summaries with desired entities provided by human.

\bibliography{anthology,custom}
\bibliographystyle{acl_natbib}

\appendix

\end{document}